\documentclass{article}

\usepackage{microtype}
\usepackage{graphicx}
\usepackage{subcaption}
\usepackage{booktabs} % for professional tables

\usepackage{hyperref}

\usepackage[preprint]{icml2026}

\usepackage{amsmath}
\usepackage{amssymb}
\usepackage{mathtools}
\usepackage{amsthm}

\usepackage[capitalize,noabbrev]{cleveref}

\theoremstyle{plain}

\theoremstyle{definition}

\theoremstyle{remark}

\usepackage[textsize=tiny]{todonotes}

\icmltitlerunning{X-VORTEX: Spatio-Temporal Contrastive Learning for Wake Vortex Trajectory Forecasting}

\begin{document}

\twocolumn[
  \icmltitle{X-VORTEX: Spatio-Temporal Contrastive Learning \\ for Wake Vortex Trajectory Forecasting}



  \icmlsetsymbol{equal}{*}

  \begin{icmlauthorlist}
    \icmlauthor{Zhan Qu}{yyy,zzz}
    \icmlauthor{Michael Färber}{yyy,zzz}
    % \icmlauthor{Firstname3 Lastname3}{comp}
  \end{icmlauthorlist}

  \icmlaffiliation{yyy}{Department of Computer Science, TU Dresden, Dresden, Germany}
  \icmlaffiliation{zzz}{ScaDS.AI, Dresden, Germany}

  \icmlcorrespondingauthor{Zhan Qu}{zhan.qu@tu-dresden.de}

  \icmlkeywords{Machine Learning, ICML}

  \vskip 0.3in
]



\printAffiliationsAndNotice{}  % no special notice (required even if empty)

\begin{abstract}
  Wake vortices are strong, coherent air turbulences created by aircraft, and they pose a major safety and capacity challenge for air traffic management. Tracking how vortices move, weaken, and dissipate over time from LiDAR measurements is still difficult because scans are sparse, vortex signatures fade as the flow breaks down under atmospheric turbulence and instabilities, and point-wise annotation is prohibitively expensive. Existing approaches largely treat each scan as an independent, fully supervised segmentation problem, which overlooks temporal structure and does not scale to the vast unlabeled archives collected in practice. We present X-VORTEX, a spatio-temporal contrastive learning framework grounded in Augmentation Overlap Theory that learns physics-aware representations from unlabeled LiDAR point cloud sequences. X-VORTEX addresses two core challenges: sensor sparsity and time-varying vortex dynamics. It constructs paired inputs from the same underlying flight event by combining a weakly perturbed sequence with a strongly augmented counterpart produced via temporal subsampling and spatial masking, encouraging the model to align representations across missing frames and partial observations. Architecturally, a time-distributed geometric encoder extracts per-scan features and a sequential aggregator models the evolving vortex state across variable-length sequences. We evaluate on a real-world dataset of over one million LiDAR scans. X-VORTEX achieves superior vortex center localization while using only 1\% of the labeled data required by supervised baselines, and the learned representations support accurate trajectory forecasting. 
\end{abstract}

\section{Introduction}

% \begin{figure}[t]
%     \centering
%     \includegraphics[width=0.9\linewidth]{figure1_example.png}
%     \caption{\textbf{The Cost of Invisible Air Turbulence.} The wreckage of American Airlines Flight 587 in Belle Harbor, NY. The crash, which resulted in 265 fatalities, was precipitated by an encounter with the wake vortex of a departing 747 \cite{NTSB_AA587}.}
%     \label{fig:motivation}
% \end{figure}

\begin{figure}[b]
    \centering
    \begin{tikzpicture}[scale=0.8]
        % elevation
        \draw[very thick, black, ->] (1.75,0) arc[start angle=0, end angle=24.624, radius=1.75];
        \node[] at (1.4,0.25)   (u) {$\varphi$};
        
        % range
        \draw[very thick, black] (0,0) -- (6,2.75);
        \draw[black,fill=black] (3.4,1.55) circle (0.06);
        \node[] at (3.2,1.8)   (s) {$R$};

        % axes
        \draw[line width=2.5,->] (0,0) -- (8,0); 
        \node[] at (7.5, -0.6)   (a) {\Large $y$};
        \draw[line width=2.5,->] (0,0) -- (0,3); 
        \node[] at (-0.6, 2.5)   (b) {\Large $z$};
        
        % give origin
        \node[] at (-0.6, 0)   (c) {\Large $z_{O}$};
        \node[] at (0, -0.6)   (d) {\Large $y_{O}$};
        
        % LiDAR
        \node[white, fill=red] at (0,0)  (h) {L};
        
        % Vortices
        % Port
        \draw[very thick, ->] (6.25,1) arc (0:140:0.3);
        \draw[very thick, ->] (6.25,1) arc (0:50:0.3);
        \draw[very thick, ->] (6.25,1) arc (0:230:0.3);
        \draw[very thick, ->] (6.25,1) arc (0:320:0.3);
        \draw[very thick] (6.25,1) arc (0:-95:0.3);
        
        \draw[black,fill=green] (5.95,1) circle (0.04);
        \node[] at (7.30,0.5)  (z) {Port vortex};
        
        \draw[black,loosely dashed,very thick] (0,1) -- (4.15,1);
        \draw[black,loosely dashed,very thick] (4.75,1) -- (5.65,1);
        \node[] at (-0.6,1)  (w) {\Large $z_{c}$};
        
        \draw[black,loosely dashed,very thick] (5.95,0.7) -- (5.95,0);
        \node[] at (5.95,-0.6)  (v) {\Large $y_{c_{\textrm{Prt}}}$};
        
        % Starboard
        \draw[very thick, ->] (4.75,1) arc (0:-320:0.3);
        \draw[very thick, ->] (4.75,1) arc (0:-230:0.3);
        \draw[very thick, ->] (4.75,1) arc (0:-140:0.3);
        \draw[very thick, ->] (4.75,1) arc (0:-50:0.3);
        \draw[very thick] (4.75,1) arc (0:95:0.3);
        
        \draw[black,fill=yellow] (4.45,1) circle (0.04);
        \node[] at (3.3,0.5)  (y) {Starboard vortex};
        
        \draw[black,loosely dashed,very thick] (4.45,0.7) -- (4.455,0);
        \node[] at (4.45,-0.6)  (r) {\Large $y_{c_{\textrm{Str}}}$};
        
        % Aircraft
        \draw[ultra thick] (4.45,1.5) .. controls (4.45,1.4) and (5.9,1.4) .. (5.95,1.5);
        \draw[very thick] (4.45,1.5) -- (4.45,1.55);
        \draw[very thick] (5.95,1.5) -- (5.95,1.55);
        \draw[thick] (5.2,1.5) -- (5.2,1.85);
        \draw[thick] (4.9,1.59) .. controls (5.15,1.555) and (5.25,1.555) .. (5.5,1.59);
        \draw[ultra thick,black,fill=white] (5.2,1.5) circle (0.115);
        \draw[very thick,black] (5.1186,1.5813) -- (5.15,1.525) -- (5.25,1.525) -- (5.2813,1.5813);
        \draw[black,fill=black] (5.2,1.445) circle (0.02);
        \draw[thick,black,fill=gray] (4.85,1.35) circle (0.07);
        \draw[black,fill=black] (4.85,1.35) circle (0.007);
        \draw[thick,black,fill=gray] (5.55,1.35) circle (0.07);
        \draw[black,fill=black] (5.55,1.35) circle (0.007);
    \end{tikzpicture}
    \caption{Measurement geometry of a ground-based LiDAR (L) scanning perpendicular to the runway. The scan captures the radial velocity of the Port and Starboard vortices trailing a landing aircraft \cite{wartha2022characterizing,qu2025explainable}.}
    \label{fig:wake-vortices-slices}
\end{figure}
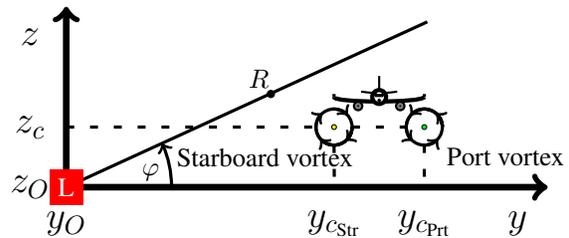

Aircraft wake vortices are coherent, counter-rotating airflows generated at the wingtips of lifting aircraft (see Fig.~\ref{fig:wake-vortices-slices}). These vortices can persist for tens of seconds and drift several hundred meters behind the generating aircraft, posing a critical safety hazard for following traffic during takeoff and landing \cite{hallock2018review, frech2002short}. The danger posed by these encounters is not merely theoretical; it has led to catastrophic loss of life and severe operational disruptions.

On November 12, 2001, American Airlines Flight 587 crashed shortly after takeoff from JFK Airport, killing all 265 people on board and 5 on the ground. The National Transportation Safety Board (NTSB) determined that the accident was precipitated by a wake vortex encounter with a preceding Boeing 747, which induced loads that snapped the Airbus A300's vertical stabilizer in mid-flight \cite{NTSB_AA587}. More recently, on January 7, 2017, a Bombardier Challenger 604 business jet operating over the Arabian Sea was caught in the wake of an Airbus A380 flying 1,000 feet above. The encounter caused the jet to roll uncontrollably and plummet 9,000 feet, resulting in extreme G-forces and a total hull loss \cite{BFU_Challenger}.

To mitigate such risks, international regulations mandate strict minimum separation distances between aircraft \cite{eurocontrol2018european}. While effective at preventing accidents, these static separations are conservative and significantly constrain airport throughput \cite{holzapfel2021mitigating}. This creates a critical trade-off between operational safety and airspace capacity. Consequently, reliable wake vortex detection and precise trajectory forecasting are essential to safely reduce separations and optimize runway efficiency, a capability that X-VORTEX aims to provide.

Recent advances in scanning LiDAR systems have enabled direct observation of wake vortices in real-world environments \cite{korner2019assessment}. Each LiDAR sweep produces a 3D point cloud of the radial velocity field (see Fig.~\ref{fig:wake-vortices-slices}), and consecutive sweeps naturally form time series that capture the evolution of wake vortex trajectories. However, existing learning-based approaches analyze these scans independently, treating each sweep as a static snapshot \cite{wartha2022characterizing,stephan2023artificial,qu2025explainable}. By relying on per-frame segmentation or detection models, prior methods largely ignore the temporal dynamics of wake evolution, such as continuous descent, decay, and dissipation driven by fluid instabilities. Moreover, supervised training of these models requires dense point-wise annotations for individual frames, which are prohibitively expensive to generate at scale \cite{smalikho2015method}.

Modeling these temporal dynamics is vital for three reasons. First, wake vortices evolve continuously through processes such as descent, deformation, and decay, so a vortex’s current geometry encodes rich information about its underlying dynamics, which is discarded by frame-wise analysis. Second, the vast majority of collected LiDAR data remains unlabeled. Leveraging temporal consistency provides a natural self-supervisory signal that can unlock the value of millions of archived scans \cite{holzapfel2021mitigating}. Third, operational wake advisory systems require more than instantaneous detection. They depend on short-term trajectory forecasting to anticipate hazardous regions and safely optimize aircraft separation, making future-state prediction central to practical deployment.

To address these challenges, we introduce \textbf{X-VORTEX} (see Fig.~\ref{fig:overview} for an overview), a spatio-temporal contrastive learning framework for wake vortex analysis. Unlike previous supervised approaches, X-VORTEX is grounded in the \textbf{Augmentation Overlap Theory} \cite{wangchaos,zhang2025augmentation}, which posits that semantically consistent augmentations induce overlapping views of the same underlying event, allowing contrastive learning to connect diverse observations and learn robust representations. We operationalize this by treating "temporal evolution" as a natural augmentation: we train the model to recognize that a spatially masked scan at time $t$ and a temporally subsampled sequence at time $t+\Delta t$ represent the same underlying physical event. X-VORTEX employs a Time-Distributed geometric encoder (e.g., PointNet) coupled with a sequential module (e.g., LSTM) to extract a physics-aware global representation from unlabeled sequences. By optimizing a multi-view InfoNCE objective, the model learns to filter out transient sensor noise while preserving the stable geometric features of the vortex cores. This pre-trained representation is then adapted to two downstream tasks: robust center localization (using a differentiable soft-center mechanism) and future trajectory forecasting.

We evaluate X-VORTEX on a massive dataset of over one million LiDAR scans collected at Vienna International Airport (LOWW) \cite{holzapfel2021mitigating}. Our experiments demonstrate that by leveraging large-scale unlabeled data, X-VORTEX significantly outperforms supervised baselines, particularly in low-data regimes. Moreover, it enables the first effective short-term forecasting of wake vortex trajectories from point cloud sequences. X-VORTEX is publicly available.\footnote{\href{https://anonymous.4open.science/r/X-Vortex-1FF4/}{https://anonymous.4open.science/r/X-Vortex-1FF4/}} In summary, the contributions of this paper are as follows:

% \begin{itemize}
%     \item \textbf{Spatio-Temporal Contrastive Framework:} We propose the first self-supervised learning pipeline for wake vortices, utilizing a time-distributed architecture to model 4D point cloud sequences.
%     \item \textbf{Physics-Driven Augmentations:} We introduce a novel data augmentation strategy based on Augmentation Overlap Theory, combining temporal subsampling with spatial masking to enforce invariance to sensor sparsity and decay rates.
%     \item \textbf{Trajectory Forecasting:} We demonstrate that our learned representations enable accurate future trajectory prediction, a capability absent in previous static segmentation models.
%     \item \textbf{Label Efficiency:} We show that pre-training on unlabeled sequences allows X-VORTEX to achieve significantly superior performance using a smaller fraction (1\%) of the labeled data required by fully supervised baselines.
% \end{itemize}

\begin{itemize}
    \item \textbf{Contrastive Pre-training for Wake Vortices:} We introduce the first self-supervised contrastive framework for aircraft generated wake vortices, learning sequence-level representations from unlabeled Doppler LiDAR data.
    \item \textbf{Dynamics-Aware View Construction:} We propose a two-view augmentation strategy combining temporal subsampling and spatial masking to enforce invariance to spatial sparsity and temporal gaps while preserving the geometric dynamics of wake vortices.
    \item \textbf{Wake Trajectory Forecasting:} We demonstrate the first short-horizon wake vortex trajectory forecasting from LiDAR point cloud sequences, outperforming trajectory-only and kinematic baselines.
    \item \textbf{Improved Label Efficiency:} We show that large-scale self-supervised pre-training substantially reduces annotation requirements, enabling accurate localization with only 1\% labeled data.
\end{itemize}

\section{Background}
\label{sec:background}

\subsection{LiDAR Remote Sensing for Aviation}
Light Detection and Ranging (LiDAR) has emerged as the gold standard for clear-air turbulence detection. Unlike radar, which relies on Rayleigh scattering from precipitation droplets (and thus suffers in clear air), Doppler LiDAR relies on Mie scattering from microscopic aerosols. This allows for high-precision measurements of wind fields even in visibly clear conditions \cite{jianbing2017survey}.

The instrument operates by emitting coherent laser pulses and analyzing the frequency shift of the back-scattered signal via the Doppler effect; this shift reveals the radial velocity of the air particles along the line of sight. By scanning a vertical slice perpendicular to the runway (Range-Height-Indicator scan), the LiDAR constructs a high-resolution 2D cross-section of the atmosphere (see Fig.~\ref{fig:wake-vortices-slices}). A sequential series of these scans effectively captures the spatio-temporal evolution of flow structures. This capacity for real-time, high-spatial-resolution velocity mapping makes LiDAR indispensable for Wake Vortex Advisory Systems, where precise localization of invisible turbulence is required to optimize aircraft separation standards safely.

\begin{figure}[tb]
 \centering 
    \includegraphics[width=0.9\linewidth]{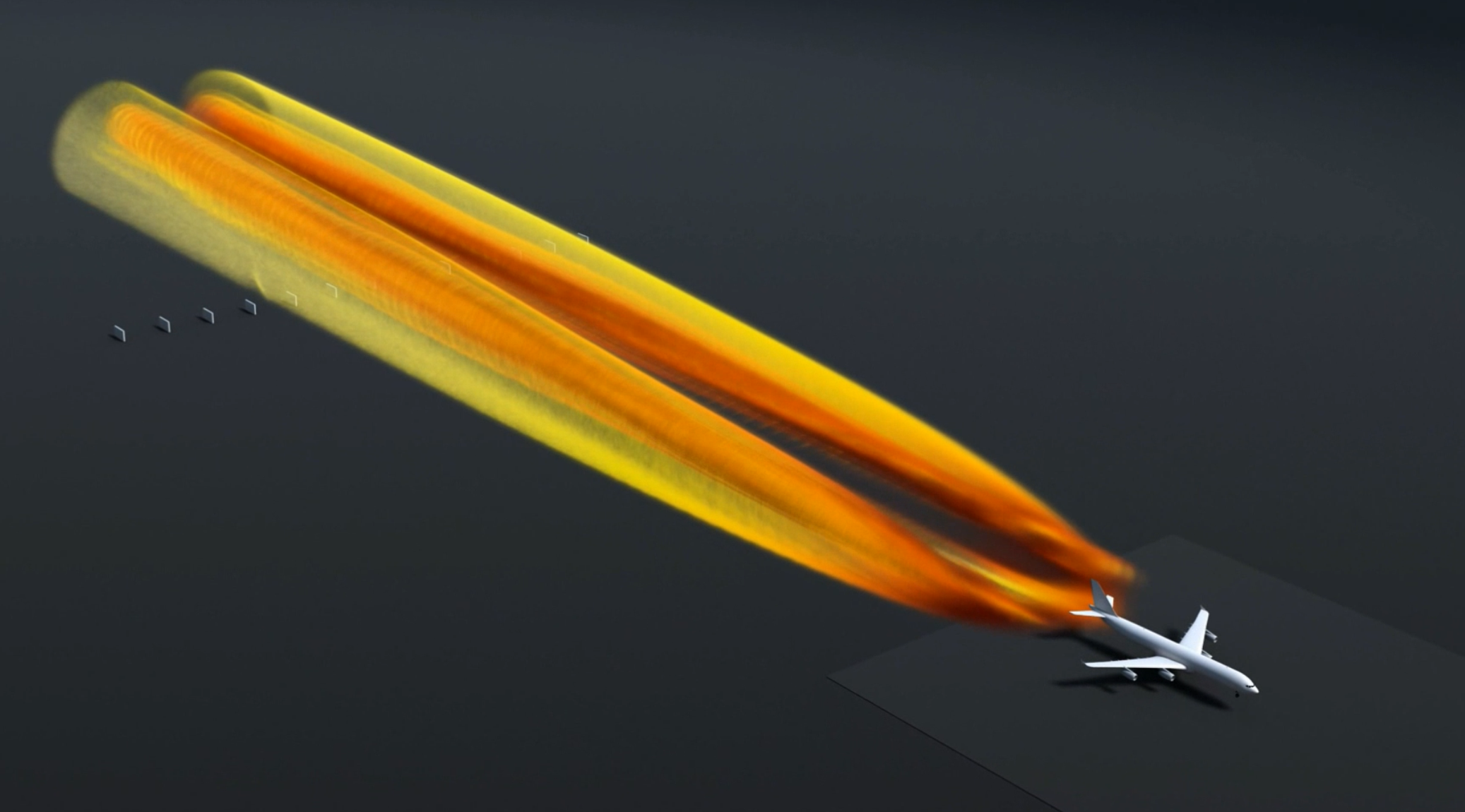}
    \caption{Large Eddy Simulation (LES) visualizing the roll-up process. Vorticity sheets shed from the wingtips and flaps merge to form the characteristic counter-rotating vortex pair.}
    \label{fig:wake-vortices-theory} 
\end{figure}

\subsection{Wake Vortex Physics and Ground Effects}
Wake vortices are an inevitable byproduct of lift generation. As an aircraft generates lift, the pressure differential between the lower (high pressure) and upper (low pressure) wing surfaces induces a spanwise flow, causing vortex sheets to shed from the wingtips and flaps \cite{stephan2019effects}. These sheets rapidly roll up into a pair of coherent, counter-rotating vortices (Fig.~\ref{fig:wake-vortices-theory}) that trail behind the aircraft for miles \cite{KUNDU2002629}.

The evolution of these vortices is a complex, multi-scale fluid dynamic problem governed by two main phases:
\begin{enumerate}
    \item \textbf{Descent and Decay:} In free air, the vortex pair descends due to mutual induction and slowly decays due to atmospheric turbulence and fluid instabilities (e.g., Crow instability) \cite{robins2001algorithm}.
    \item \textbf{Ground Interaction:} During landing, vortices descend into the ground effect zone. This boundary interaction causes the trajectory to diverge hyperbolically \cite{harvey1971flowfield}. Furthermore, the no-slip condition at the runway surface generates secondary vorticity, which can cause the primary vortices to rebound upward or hover over the runway \cite{spalart2001modeling,zheng1996study,holzapfel2007aircraft}.
\end{enumerate}
This non-linear interaction with the ground, combined with crosswind variability, makes analytical prediction of vortex trajectories notoriously difficult. Consequently, data-driven approaches that can model these spatio-temporal dynamics from measurement data are highly desirable.

\section{Related Work}
\subsection{Wake Vortex Analysis and Forecasting} 

Early data-driven approaches to wake vortex analysis applied traditional machine learning methods, such as Support Vector Machines (SVMs), random forests, and clustering algorithms to handcrafted features extracted from LiDAR scans \cite{pan2020identification,weijun2021aircraft,pan2022recognition,felton2025characterizing}. While these methods demonstrated the feasibility of automated wake vortex detection, they relied on superficial scan characteristics and lacked the capacity to represent the complex, multiscale structure of wake vortices. With the advent of deep learning, LiDAR scans were increasingly converted into 2D images and processed using Convolutional Neural Networks (CNNs), Long Short-Term Memory (LSTM) networks, or hybrid CNN--LSTM architectures for wake vortex detection, segmentation, and localization \cite{abras2021application,ai2021deep,chu2024assessment,wartha2022characterizing,zhang2024locating,baranov2021wake}. More recently, advanced computer vision models such as YOLO and VGG have further improved detection performance on image-based representations of LiDAR data \cite{weijun2019deep,pan2022identification,stietz2022artificial,stephan2023artificial}.

Despite their success, image-based pipelines require substantial preprocessing and inherently discard the native geometric structure of LiDAR measurements, limiting physical interpretability and explainability \cite{stephan2023artificial,wartha2022characterizing}. To address this limitation, recent work has begun to operate directly on LiDAR-derived 3D point clouds, which preserve spatial relationships and allow models to exploit the full structure of the measurements. Point cloud--specific architectures such as PointNet \cite{qi2017pointnet}, PointNet++ \cite{qi2017pointnet++}, and Dynamic Graph Convolutional Neural Networks (DGCNN) \cite{DGCNN_WSLSBS19} have demonstrated strong performance in segmentation and classification tasks across multiple domains \cite{Survey3D_GWHLLB21}. Building on this paradigm, prior work introduced the first application of 3D point cloud semantic segmentation for wake vortex detection, leveraging DGCNN-based models combined with clustering and perturbation-based explanations to achieve accurate and interpretable vortex localization directly from LiDAR scans \cite{qu2025explainable}.

However, existing learning-based approaches, both image-based and point cloud-based, remain fundamentally static, processing each LiDAR scan independently and ignoring the strong temporal coherence of wake vortex evolution. As a result, these methods are limited to instantaneous detection or localization and do not explicitly model wake vortex trajectories. This represents a critical gap for operational Wake Vortex Advisory Systems, where decision-making depends not only on identifying current vortices but also on forecasting their future motion and dissipation to safely optimize aircraft separation standards. Moreover, current approaches rely exclusively on supervised learning and require densely annotated scans, which are expensive to obtain and severely limit scalability to large operational datasets.

\subsection{Deep Learning on 4D Point Clouds}

Learning from sequences of point clouds (i.e., 4D data with time as an explicit dimension) has emerged as an active research area, primarily driven by applications in autonomous driving, robotics, and human motion analysis. Early approaches extended static point cloud encoders by aggregating temporal information using Recurrent Neural Networks (RNNs), enabling sequential feature propagation across frames \cite{liu2019meteornet,fan2019pointrnn}. While effective for short temporal contexts, these methods often struggle to model long-range temporal dependencies and complex spatio-temporal interactions.

More recent architectures explicitly model space and time jointly. Sparse 4D convolutional networks, such as Minkowski-based 4D ConvNets, extend voxel-based convolutions to the spatio-temporal domain and enable efficient learning on large-scale point cloud sequences \cite{choy20194d}. Point-based methods have also introduced spatio-temporal convolution operators to directly capture local motion patterns across consecutive frames \cite{fan2022pstnet}. In parallel, Transformer-based models leverage self-attention mechanisms to model long-range dependencies in 4D point clouds, achieving strong performance on dynamic scene understanding tasks \cite{fan2021point}. These approaches have further been extended to temporally consistent semantic and panoptic segmentation of LiDAR sequences \cite{aygun20214d}.

Despite their expressive power, existing 4D point cloud models are largely designed for scenarios involving rigid or piecewise-rigid motion, such as vehicles, pedestrians, or articulated human bodies, where object identity and geometric structure are preserved over time \cite{aygun20214d,choy20194d}. In contrast, aircraft wake vortices constitute non-rigid, transient fluid structures whose geometry continuously deforms, diffuses, and dissipates under atmospheric conditions. This fundamental difference limits the direct applicability of existing 4D point cloud methods. Our work addresses this gap by adapting 4D representation learning to wake vortex dynamics, employing a time-distributed architecture that explicitly separates spatial vortex geometry from temporal evolution and decay processes, enabling both detection and trajectory forecasting.

\begin{figure*}[!ht]
    \centering
    \includegraphics[width=\linewidth]{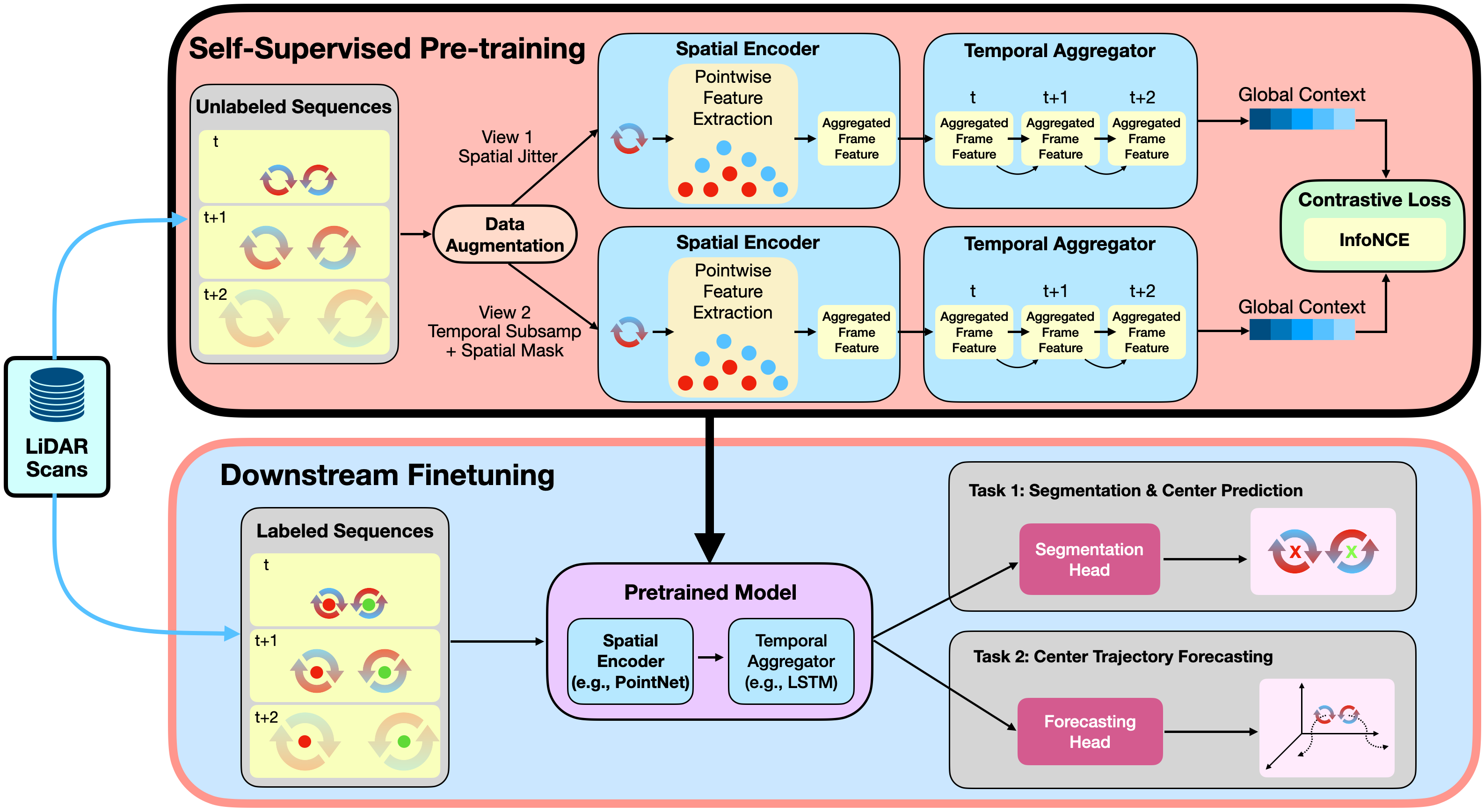}
    \caption{\textbf{Overview of X-VORTEX.}
    Top: self-supervised pre-training on unlabeled wake sequences using weak and strong augmentations of the same sequence, optimized with InfoNCE. Bottom: downstream adaptation on labeled sequences, where the pretrained spatial encoder and temporal aggregator are frozen to produce representations for (i) vortex center localization via a segmentation-based soft-center head and (ii) short-horizon trajectory forecasting via a lightweight prediction head.}
    \label{fig:overview}
\end{figure*}

\subsection{Self-Supervised Representation Learning}

Self-supervised learning (SSL), and contrastive learning in particular, has become a dominant paradigm for representation learning by enabling models to extract semantic structure from large-scale unlabeled data \cite{chen2020simple,he2020momentum}. From a theoretical perspective, contrastive learning has been shown to promote both alignment between positive pairs and uniformity of representations over the embedding space \cite{wang2020understanding}. In the 3D domain, contrastive approaches such as PointContrast \cite{xie2020pointcontrast} and DepthContrast \cite{zhang2021self} learn invariant representations by enforcing consistency across augmented views of the same underlying geometry, typically at the point or scene level. While effective, most existing 3D SSL methods focus on static scenes, where two views correspond to the same geometry under different spatial transformations.

Extending contrastive learning to temporally evolving data introduces additional challenges, as semantic consistency must be maintained despite genuine structural change over time. Naively contrasting temporally adjacent frames risks suppressing meaningful dynamics, while treating distant frames as negatives may violate underlying continuity. Recent theoretical work on \emph{augmentation overlap} \cite{wangchaos,zhang2025augmentation} explains how strong but semantically consistent augmentations generate overlapping views of the same sample, enabling contrastive learning to capture shared structure across transformations without collapsing meaningful variation.

Building on these insights, X-VORTEX adopts a spatio-temporal contrastive learning objective tailored to fluid-dynamic point cloud sequences. By treating temporally non-adjacent observations of the same wake vortex and spatially masked regions as positive pairs, the model is encouraged to learn representations that are invariant to decay and deformation while remaining sensitive to physically meaningful evolution. This enables self-supervised learning of wake vortex dynamics without requiring dense temporal annotations.

\section{Methodology}
\label{sec:method}

\subsection{Problem Formulation}
We formulate the wake vortex analysis problem as learning from a sequence of unstructured 3D point clouds $\mathcal{S} = \{\mathbf{P}_t\}_{t=1}^{T}$, where each frame $\mathbf{P}_t \in \mathbb{R}^{N \times 3}$ represents a LiDAR scan at time $t$, normalized to a fixed number of points $N$. The sequence captures the temporal evolution of a vortex pair (port and starboard) generated by an aircraft. Unlike traditional static segmentation, we aim to model the spatio-temporal dynamics of vortex evolution. Our objective is twofold:
\begin{enumerate}
    \item \textbf{Unsupervised Representation Learning:} Learn a physics-aware embedding $Z_\mathcal{S}$ from a large corpus of unlabeled sequences by exploiting temporal continuity and spatial invariance.
    \item \textbf{Downstream Detection \& Forecasting:} Utilize the learned representation to (a) robustly predict the vortex center coordinates $(y_c, z_c)$ at time $t$, and (b) forecast the future trajectory at $t+1, \dots, t+k$.
\end{enumerate}

\subsection{The X-VORTEX Framework}
We propose \textbf{X-VORTEX} (see Fig.~\ref{fig:overview}), a contrastive learning framework grounded in Augmentation Overlap Theory~\cite{wangchaos,zhang2025augmentation}. The theory shows that contrastive learning is effective when augmented views of the same sample retain sufficient distributional overlap while remaining separable from other samples. X-VORTEX leverages this principle to learn robust representations from unlabeled wake vortex sequences. The architecture consists of three core components:

\paragraph{1. Spatio-Temporal Encoder.} To process 4D data efficiently, we employ a time-distributed architecture:
\begin{itemize}
    \item \textbf{Spatial Backbone (e.g., PointNet):} Each frame $\mathbf{P}_t$ in the sequence is independently processed by a shared PointNet++ encoder $f_\theta$, yielding a sequence of spatial feature vectors $\{\mathbf{v}_1, \dots, \mathbf{v}_T\}$, where $\mathbf{v}_t \in \mathbb{R}^{D_s}$.
    \item \textbf{Temporal Aggregator (e.g., LSTM):} The sequence of spatial features is fed into a Long Short-Term Memory (LSTM) network $g_\phi$ to capture the vortex decay dynamics. The final hidden state $\mathbf{h}_T \in \mathbb{R}^{D_t}$ serves as the global context vector for the entire sequence.
\end{itemize}

\paragraph{2. Augmentation Strategy (The "Two Views").} We construct positive pairs by generating two distinct views of the same underlying flight event, specifically designed to enforce physics-based constraints:
\begin{itemize}
    \item \textbf{View 1 (Weak Augmentation):} The original sequence with minor spatial jitter. This preserves the exact temporal structure.
    \item \textbf{View 2 (Strong Augmentation):} A perturbed version constructed via:
    \begin{itemize}
        \item \textit{Temporal Subsampling:} Frames are randomly skipped (e.g., $t, t+2, t+4$) to force the model to infer missing intermediate states.
        \item \textit{Spatial Masking:} Random point dropout (20-30\%) and rigid rotation around the vertical axis ($z$). This forces the encoder to rely on global geometry rather than local sensor artifacts.
    \end{itemize}
\end{itemize}

\paragraph{3. Contrastive Objective.} We employ the InfoNCE loss to align representations of two augmented views of the same sequence. For each sample, a weakly perturbed anchor view and a strongly augmented counterpart (temporal subsampling and spatial masking) form a positive pair, while views from other sequences in the batch act as negatives. For a batch of $B$ sequences, the loss for the $i$-th sample is:
$$\mathcal{L}_i = -\log \frac{\exp(\text{sim}(\mathbf{h}_i^{(1)}, \mathbf{h}_i^{(2)}) / \tau)}{\sum_{j=1}^{2B} \mathbb{1}_{[j \neq i]} \exp(\text{sim}(\mathbf{h}_i^{(1)}, \mathbf{h}_j) / \tau)}$$
where $\mathbf{h}^{(1)}$ and $\mathbf{h}^{(2)}$ are the projected embeddings of the two views, and $\tau$ is the temperature parameter. This objective pulls the strongly perturbed view toward its corresponding anchor, encouraging the model to learn representations that are invariant to spatial sparsity and temporal gaps while preserving the underlying wake dynamics. The weak view retains fine-grained geometric structure, whereas the strong view forces the model to infer wake evolution from incomplete spatial and temporal observations, mirroring real-world LiDAR sensing conditions.

\subsection{Downstream Task Adaptation}
After pre-training on unlabeled data, we freeze the spatial backbone and fine-tune lightweight heads on the labeled dataset:

\textit{Task 1: Context-Aware Center Prediction.} We introduce a Soft-Center Segmentation Head. Instead of direct regression, we predict a "vortexness" score map $M \in [0,1]^N$ for the points in the final frame. The center is computed as the differentiably weighted center of mass:
$$\mathbf{c}_{pred} = \frac{\sum_{i=1}^N M_i \cdot \mathbf{p}_i}{\sum_{i=1}^N M_i}$$
This allows the supervision signal (MSE on center coordinates) to backpropagate directly into the segmentation mask, refining the spatial localization.

% \textit{Task 2: Trajectory Forecasting.} A Multi-Layer Perceptron (MLP) takes the global context vector $\mathbf{h}_T$ and regresses the center coordinates for future time steps $t+1$ and $t+2$.

\textit{Task 2: Trajectory Forecasting.} A Multi-Layer Perceptron (MLP) takes the global context vector $\mathbf{h}_T$, computed from a brief history of the three most recent LiDAR scans, and regresses the vortex center coordinates for future time steps $t+1$ and $t+2$.

\section{Experimental Setup}
\label{sec:experiments}

\subsection{Dataset and Pre-processing} 
\label{sec:experiments_data}
We utilize a large-scale LiDAR dataset of aircraft wake vortices collected at Vienna International Airport, comprising a total of 1,032,694 raw scans \cite{holzapfel2021mitigating}.

\textbf{Labeled Subset.} The labeled portion consists of 1,176 flight sequences (19,354 scans) with annotated vortex centers, where sequence lengths range from 1 to 52 scans. To ensure consistent temporal context, we retain only sequences with at least five scans; longer sequences are either split into multiple segments or truncated to a fixed length of $T{=}5$. This yields 3,425 sequences containing 17,125 scans in total. The data are split into train/validation/test sets with 2,398/685/342 sequences (11,990/3,425/1,710 scans), respectively.

Ground-truth annotations are provided only for vortex center locations and are generated using a physics-based wake vortex characterization algorithm, the Radial Velocity (RV) method \cite{smalikho2015method}. These labels are subject to inherent uncertainty, particularly under turbulent atmospheric conditions, and recent studies have quantified non-negligible localization errors in such scenarios \cite{wartha2023investigating}. We therefore regard these annotations as approximate, algorithm-generated labels rather than exact ground truth.

\textbf{Unlabeled Subset (for Pre-training).} For self-supervised pre-training, we construct an unlabeled dataset of 194,864 sequences (974,320 scans) by grouping consecutive scans with $\Delta t < 8.0\,\mathrm{s}$ and filtering for valid point densities (at least 8,000 points per scan) to ensure data quality. This subset is split into train/validation/test partitions using a 70\%/20\%/10\% ratio, resulting in 136,406/38,972/19,486 sequences (682,030/194,860/97,430 scans), respectively.

\textit{Sequence Construction.}
To strictly prevent data leakage and false negatives in contrastive learning, we employ a Non-Overlapping Chunking strategy. Long flight recordings are split into fixed-length segments of $T{=}5$ frames with no overlap. We apply \emph{sequence-level centering} to remove global position and altitude offsets: for each sequence $\mathcal{S}=\{\mathbf{P}_t\}_{t=1}^{T}$, we compute a single centroid over all points across all frames,
$\boldsymbol{\mu}_{\mathcal{S}} = \frac{1}{TN}\sum_{t=1}^{T}\sum_{i=1}^{N}\mathbf{p}_{t,i}$,
and translate every point as $\mathbf{p}'_{t,i}=\mathbf{p}_{t,i}-\boldsymbol{\mu}_{\mathcal{S}}$. This normalization preserves relative geometry and motion within the sequence while preventing the model from exploiting absolute height or site-specific coordinate offsets.

\subsection{Baselines}
We compare X-VORTEX against representative baselines covering heuristic methods, image-based pipelines, static 3D point cloud models, and temporal models without self-supervised pre-training.

\textit{Heuristic Baselines.}
\textbf{DBSCAN Clustering} estimates vortex centers as centroids of density-based clusters, while \textbf{Intensity Centroid} computes weighted centroids of high-magnitude radial velocity returns. These methods require no learning.

\textit{2D Image-Based Baselines.}
LiDAR scans are projected into 2D images and processed using a \textbf{Standard CNN} and \textbf{YOLOv8}. Predicted vortex regions are converted to center coordinates in physical space.

\textit{Static 3D Point Cloud Baselines.}
\textbf{PointNet}, \textbf{PointNet++}, and \textbf{DGCNN} operate directly on 3D point clouds but process each scan independently, without temporal context.

\textit{Temporal Baselines.} A \textbf{Supervised LSTM} combines a time-distributed PointNet++ encoder with an LSTM and is trained end-to-end using labeled data only. 

\textit{Trajectory Forecasting Baselines.} A \textbf{Trajectory-Only LSTM} predicts future vortex centers using past center coordinates, without access to point cloud geometry. We further include two kinematic baselines, \textbf{Constant Velocity} and a \textbf{Kalman Filter}, which extrapolate future positions based on simple motion assumptions.

\subsection{Implementation Details}
\label{sec:impl_details}

\textit{Architecture.}
Unless otherwise stated, X-VORTEX employs PointNet as the spatial encoder and a single-layer LSTM with hidden dimension 256 as the temporal aggregator. The LSTM processes per-frame embeddings and produces a sequence-level representation, which is passed through a two-layer MLP projection head during contrastive pre-training. 

% For downstream evaluation, we freeze the spatial encoder and train lightweight task-specific heads for vortex center localization and short-horizon trajectory prediction.

\textit{Pre-training protocol.}
We pre-train on the unlabeled subset for up to 100 epochs with a global batch size of 64 sequences (128 augmented views per step) using Adam with an initial learning rate of $10^{-3}$ and a cosine decay schedule. Training is performed on 4 NVIDIA H100 GPUs, with wall-clock time ranging from approximately 40 to 72 hours depending on the backbone architecture. In practice, models typically converge before the full 100 epochs (around 80 epochs). We use the InfoNCE objective with temperature $\tau{=}0.07$ and standard in-batch negatives. To avoid false negatives due to scan overlap across training examples, we employ the non-overlapping chunking strategy described in Sec.~\ref{sec:experiments_data}.

\textit{Augmentations.}
We follow the two-view design in Sec.~\ref{sec:method}. View~1 applies weak perturbations (minor spatial jitter). View~2 applies strong augmentations: temporal subsampling (random frame skipping within each $T{=}5$ chunk), spatial masking via random point dropout with ratio $p{=}0.3$, and rigid rotation around the vertical axis with range $\pm 30^\circ$. These parameters are fixed across all experiments for fair comparison.

\textit{Downstream fine-tuning.}
For center localization, we freeze the pre-trained spatial encoder and fine-tune the temporal module together with task-specific prediction heads on the labeled subset. For trajectory forecasting, we train an MLP regressor on top of the sequence-level embedding using the same labeled split.

\subsection{Evaluation Metrics}
\label{sec:metrics}

\textit{Center Localization Error.}
We measure center localization quality using Euclidean RMSE (in meters) between predicted and ground-truth vortex centers in the final frame of each sequence. This metric directly reflects operational localization accuracy.

\textit{Forecasting Error.}
For trajectory forecasting, we report RMSE (meters) at horizons $t{+}1$ and $t{+}2$, corresponding to approximately 6-8\,s and 12-16\,s ahead in our sequence sampling.

\textit{Representation Quality and Geometry.}
We evaluate unsupervised representation quality via linear probing accuracy on the labeled set. To diagnose contrastive training behavior, we also report alignment and uniformity metrics to quantify how well positive pairs are pulled together while avoiding embedding collapse.

\section{Results}
\label{sec:results}

We evaluate X-VORTEX along three dimensions: (i) representation quality learned from unlabeled sequences, (ii) label efficiency for center localization under scarce supervision, and (iii) the ability to forecast future wake trajectories. We further provide an ablation study to isolate the contributions of key design choices, and qualitative evidence to illustrate typical failure modes and robustness.

\begin{table}[!htb]
    \centering
    \caption{\textbf{Linear Probing Accuracy.}
    Classification accuracy (\%) of a linear classifier trained on frozen representations for aircraft type classification (16 classes).}
    \label{tab:linear_probing}
    \resizebox{\linewidth}{!}{%
    \begin{tabular}{lllc}
        \toprule
        \textbf{Method} & \textbf{Backbone} & \textbf{Input View} & \textbf{Acc (\%)} \\
        \midrule
        Random Initialization & PointNet++ & Frame & 14.21 \\
        Spatial-Only & PointNet++ & Frame & 29.39 \\
        \midrule
        \textbf{X-VORTEX (Ours)} & \textbf{PointNet} & Sequence & \textbf{72.03} \\
        X-VORTEX (Ours) & PointNet++ & Sequence & 69.41 \\
        X-VORTEX (Ours) & DGCNN & Sequence & 67.02 \\
        \bottomrule
    \end{tabular}
    }
\end{table}

\subsection{Unsupervised Representation Quality}
\label{sec:repr_quality}

We evaluate representation quality using a standard \textbf{linear probing} protocol: the encoder is frozen and a linear classifier is trained on top to predict aircraft type (16 classes, see Appendix~\ref{app:probing} for details) from the labeled subset. While aircraft classification is not a downstream task of interest per se, it provides a diagnostic measure of how linearly separable different wake patterns become under the learned embedding.

As shown in Table~\ref{tab:linear_probing}, X-VORTEX substantially improves linear separability compared to random initialization and spatial-only baselines. In particular, sequence-level representations learned with X-VORTEX achieve up to 72.03\% accuracy, whereas single-frame PointNet++ features trained without temporal context reach only 29.39\%. This gap indicates that incorporating temporal information during self-supervised pre-training yields embeddings that better represent wake-induced flow structures.

To further analyze representation geometry beyond linear probing, we report alignment and uniformity metrics in Appendix~\ref{sec:align_uniform}, providing complementary evidence of stable contrastive learning and well-distributed embeddings.

\begin{table}[t]
    \centering
    \caption{\textbf{Center Localization Error.}
    RMSE (meters) of predicted wake vortex centers under different fractions of labeled training data. $\Delta$~1\% denotes the relative improvement of X-VORTEX over the best baseline at 1\% labeled data.}
    \label{tab:segmentation_results}
    \resizebox{\linewidth}{!}{%
    \begin{tabular}{l|ccc|c}
        \toprule
        & \multicolumn{3}{c|}{\textbf{Labeled Data Fraction}} & \\
        \textbf{Method} & \textbf{1\%} & \textbf{10\%} & \textbf{100\%} & \textbf{\boldmath$\Delta$ 1\%} \\
        \midrule
        \textit{Heuristics (No Learning)} & & & & \\
        DBSCAN Clustering & 117.45 m & 117.45 m & 117.45 m & -- \\
        Intensity Centroid & 118.77 m & 118.77 m & 118.77 m & -- \\
        \midrule
        \textit{2D Image Baselines (Projected)} & & & & \\
        Standard CNN & 56.78 m & 35.82 m & 30.10 m & -- \\
        YOLOv8 & 27.61 m & 25.23 m & 22.44 m & -- \\
        \midrule
        \textit{Static 3D Baselines (Supervised)} & & & & \\
        PointNet (Single Frame) & 43.75 m & 24.54 m & 10.18 m & -- \\
        PointNet++ (Single Frame) & 42.18 m & 33.33 m & 11.93 m & -- \\
        DGCNN (Single Frame) & 49.13 m & 29.19 m & 6.33 m & -- \\
        \midrule
        \textit{Spatio-Temporal Models (4D)} & & & & \\
        Supervised PointNet + LSTM & 53.14 m & 34.60 m & 17.19 m & -- \\
        \midrule
        \textit{X-VORTEX (finetuned)} & & & & \\
        \textbf{X-VORTEX (PointNet)} & \textbf{9.15 m} & \textbf{7.48 m} & \textbf{5.15 m} & \textbf{+66.8\%} \\
        X-VORTEX (PointNet++) & 15.71 m & 10.62 m & 7.37 m & +43.1\% \\
        X-VORTEX (DGCNN) & 20.02 m & 17.88 m & 10.19 m & +27.5\% \\
        \bottomrule
    \end{tabular}
    }
\end{table}

\subsection{Label Efficiency in Center Localization}
\label{sec:label_efficiency}

We evaluate vortex center localization using RMSE (meters) under varying fractions of labeled training data (1\%, 10\%, and 100\%). Results are reported in Table~\ref{tab:segmentation_results}. For heuristic methods, which do not depend on labeled training data, the same test-set error is reported across all label fractions. The $\Delta$~1\% column denotes the relative improvement of each X-VORTEX variant over the strongest non-self-supervised baseline (YOLOv8) at 1\% labeled data.

Across all settings, X-VORTEX consistently outperforms heuristic, image-based, and fully supervised point-cloud baselines. In the extreme low-label regime (1\%), X-VORTEX with a PointNet backbone achieves 9.15\,m RMSE, corresponding to a 66.8\% reduction in error relative to YOLOv8. Even with alternative backbones, X-VORTEX maintains substantial gains, improving over the best baseline by 43.1\% (PointNet++) and 27.5\% (DGCNN), demonstrating that the proposed spatio-temporal pre-training strategy generalizes across architectures.

Static 3D models trained from scratch degrade severely under limited supervision, with errors exceeding 40\,m at 1\% labels, highlighting the difficulty of learning reliable wake structure from sparse annotations alone. In contrast, X-VORTEX leverages large-scale unlabeled sequences to learn transferable representations prior to fine-tuning, enabling accurate localization with minimal labeled data. As label availability increases, X-VORTEX continues to outperform competing methods, reaching 5.15\,m RMSE at full supervision.

We observe that the PointNet backbone consistently outperforms PointNet++ and DGCNN across localization and forecasting tasks. This behavior aligns with the physical characteristics of wake vortices, which manifest as diffuse, non-rigid flow structures whose discriminative cues are primarily global (e.g., large-scale velocity dipoles and circulation patterns), rather than localized geometric primitives. PointNet’s global pooling therefore better captures these vortex-scale signatures, while neighborhood-based models rely on local graph constructions that become unstable under LiDAR sparsity, spatial masking, and temporal deformation. This effect is further amplified by our contrastive augmentations, which perturb local neighborhoods while preserving global structure, making simpler global encoders more robust in the self-supervised setting.

\textit{Discussion.}
These results indicate that incorporating temporal context through self-supervised pre-training provides a strong initialization for wake vortex localization. By exploiting unlabeled sequence data rather than relying solely on labeled scans, X-VORTEX substantially reduces annotation requirements while maintaining competitive performance under full supervision.

\subsection{Wake Vortex Trajectory Forecasting}
\label{sec:forecasting}

% Beyond localizing current wake positions, operational systems benefit from anticipating where hazardous wakes will move next. We evaluate short-horizon forecasting at $t{+}1$ and $t{+}2$ using a brief history of LiDAR scans.

Beyond localizing current wake positions, operational systems benefit from anticipating where hazardous wakes will move next. We evaluate short-horizon forecasting at $t{+}1$ and $t{+}2$ using a brief history of 3 past scans, which provides minimal temporal context while reflecting realistic sensing constraints in operational settings.

Results are summarized in Table~\ref{tab:forecasting_results}. Kinematic baselines based on Constant Velocity and Kalman filtering exhibit large errors at both horizons, reflecting their inability to capture the non-linear descent and decay of wake vortices. A Trajectory-Only LSTM, which operates solely on past center coordinates, provides only marginal improvement over these methods, indicating that center trajectories alone are insufficient for reliable prediction.

In contrast, X-VORTEX substantially reduces forecasting error across all backbones by leveraging sequence-level representations learned from full point cloud inputs. The PointNet variant achieves the lowest RMSE, with 19.99\,m at $t{+}1$ and 22.06\,m at $t{+}2$, while PointNet++ and DGCNN yield slightly higher but still competitive performance. The growing performance gap at the longer horizon highlights the benefit of incorporating spatial structure when extrapolating wake motion under increasing uncertainty.

\textit{Discussion.}
These results indicate that effective wake forecasting requires more than modeling past center trajectories or applying simple kinematic assumptions. By combining temporal context with point-cloud geometry through self-supervised pre-training, X-VORTEX learns representations that support more stable short-term prediction, providing a stronger foundation for forecasting than coordinate-only approaches.

\begin{table}[!ht]
    \centering
    \caption{\textbf{Trajectory Forecasting Error.}
    RMSE (meters) for predicting wake vortex center positions at future time horizons $t{+}1$ and $t{+}2$.}
    \label{tab:forecasting_results}
    \resizebox{\linewidth}{!}{%
    \begin{tabular}{lcc}
        \toprule
        \textbf{Method} & \textbf{Horizon $t+1$} & \textbf{Horizon $t+2$} \\
        \midrule
        Constant Velocity & 54.56 m & 58.27 m \\
        Kalman Filter & 53.83 m & 56.57 m \\
        Trajectory-Only LSTM & 53.11 m & 55.23 m \\
        \midrule
        \textbf{X-VORTEX (PointNet)} & \textbf{19.99 m} & \textbf{22.06 m} \\
        X-VORTEX (PointNet++) & 22.31 m & 24.76 m \\
        X-VORTEX (DGCNN) & 25.37 m & 32.79 m \\
        \bottomrule
    \end{tabular}
    }
\end{table}

\subsection{Ablation Studies}
\label{sec:ablation}

We ablate key components of X-VORTEX to assess their contribution to both center localization and trajectory forecasting. Results are reported in Table~\ref{tab:ablation_study}.

\begin{table}[!htb]
    \centering
    \caption{\textbf{Ablation Results.}
    RMSE for center localization and trajectory forecasting at horizon $t{+}1$ (meters). $\Delta$ denotes absolute degradation relative to the full X-VORTEX model.}
    \label{tab:ablation_study}
    \resizebox{0.95\linewidth}{!}{%
    \begin{tabular}{lcc}
        \toprule
        \textbf{Configuration} & \textbf{Center ($\Delta$)} & \textbf{Horizon $t+1$ ($\Delta$)} \\
        \midrule
        \textbf{X-VORTEX (PointNet)} & \textbf{5.15 m} & \textbf{19.99 m} \\
        \midrule
        \textit{Augmentations} & & \\
        w/o Temporal Subsampling (View 2) & 18.98 m (+13.83) & 29.92 m (+9.93) \\
        w/o Spatial Masking (View 2) & 16.99 m (+11.84) & 28.84 m (+8.85) \\
        \midrule
        \textit{Normalization \& Architecture} & & \\
        w/o Altitude Centering (Raw coords) & 10.91 m (+5.76) & 25.50 m (+5.51) \\
        w/o LSTM (Mean Pooling) & 30.42 m (+25.27) & 52.79 m (+32.80) \\
        \bottomrule
    \end{tabular}
    }
\end{table}

\textit{Temporal subsampling is essential.}
Removing temporal subsampling substantially degrades performance, increasing center localization error by 13.83\,m and forecasting error at $t{+}1$ by 9.93\,m. This confirms that forcing the model to bridge temporal gaps during pre-training is critical for learning representations that support both localization and prediction.

\textit{Spatial masking improves robustness.}
Disabling spatial masking leads to consistent performance drops (+11.84\,m for localization and +8.85\,m for forecasting), indicating that invariance to point sparsity and missing returns plays an important role in stabilizing downstream predictions.

\textit{Sequence-level centering prevents shortcut learning.}
Removing altitude centering increases localization and forecasting errors by 5.76\,m and 5.51\,m, respectively, suggesting that normalization is necessary to prevent reliance on absolute scan placement and to encourage learning of relative wake structure.

\textit{Explicit temporal modeling matters beyond pooling.}
Replacing the LSTM with mean pooling causes the largest degradation (+25.27\,m for localization and +32.80\,m for forecasting), demonstrating that ordered temporal modeling is essential and cannot be replaced by unordered aggregation of frame-level features.

\begin{figure}[!ht]
    \centering
    \includegraphics[width=0.5\textwidth]{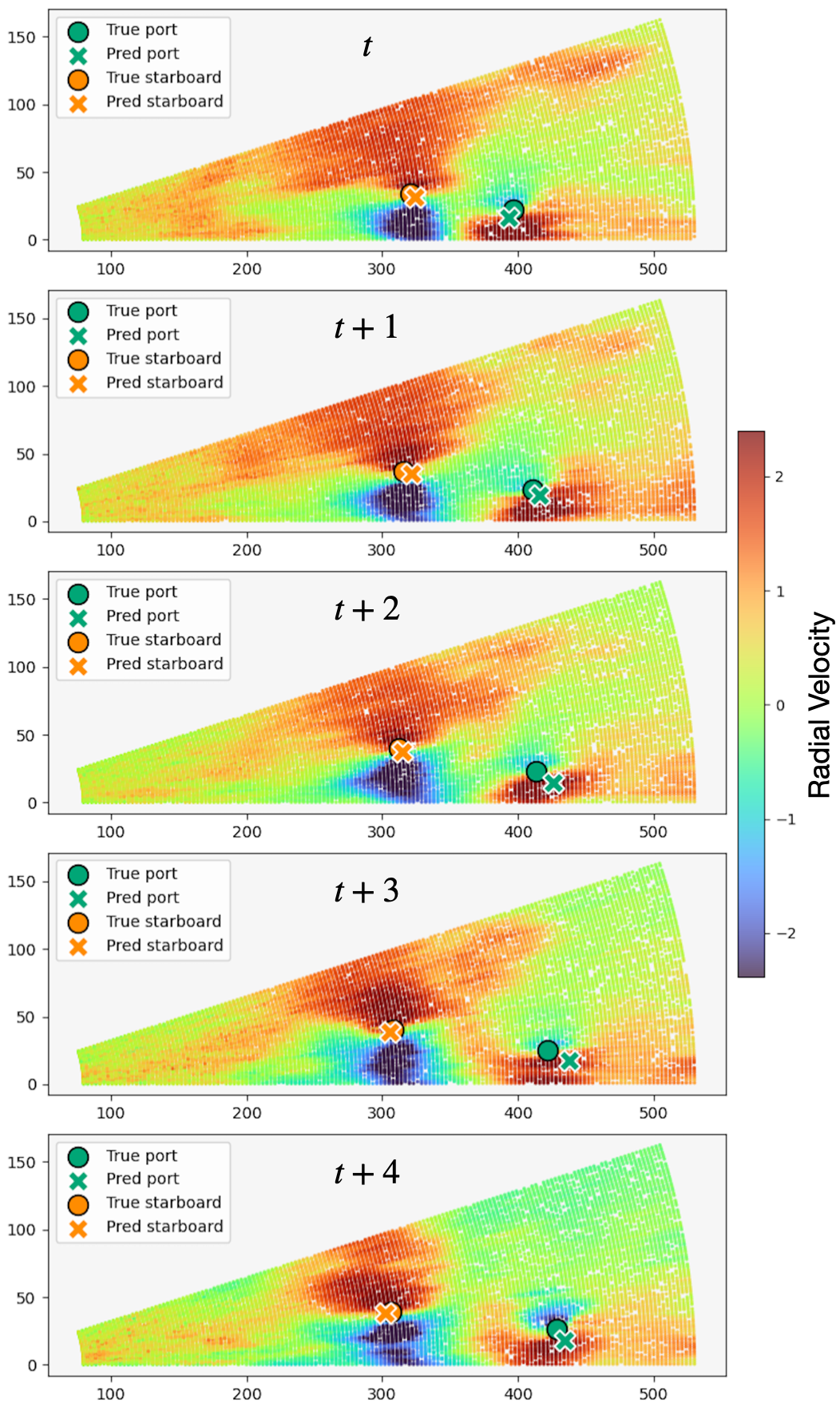}
    \caption{\textbf{Qualitative Visualization.} A sequence of five consecutive LiDAR scans ($t$ to $t{+}4$) showing the radial velocity field, where warm colors indicate motion toward the sensor and cool colors indicate motion away. The counter-rotating port and starboard vortices appear as paired regions of opposite velocity. Green and orange circles denote ground-truth vortex centers, while crosses indicate model predictions.}
    \label{fig:qualitative}
\end{figure}

\subsection{Qualitative Results}
\label{sec:qualitative_results}

Figure~\ref{fig:qualitative} shows a representative sequence of five consecutive LiDAR scans ($t$ to $t{+}4$) together with ground-truth and predicted vortex centers. Each panel visualizes the radial velocity field measured by the Doppler LiDAR, where warm colors (red/orange) indicate motion toward the sensor and cool colors (blue) indicate motion away from the sensor. The wake consists of a counter-rotating vortex pair, and each vortex therefore exhibits adjacent regions of positive and negative radial velocity, corresponding to the port and starboard vortices trailing the aircraft. Due to their opposite rotation directions, the two vortices display complementary velocity patterns: one typically shows stronger toward-sensor motion in its upper region and away-sensor motion below, while the other exhibits the inverse structure. These characteristic signatures provide important geometric cues for localization but become progressively weaker and more diffuse as the wake decays over time. Green and orange circles denote ground-truth centers for the port and starboard vortices, respectively, while crosses indicate model predictions. 

Despite this progressive decay and increasing sparsity, X-VORTEX maintains stable center predictions across the sequence, closely tracking both vortices even when wake signatures are faint. This highlights the benefit of jointly leveraging temporal context and spatial structure: rather than relying on isolated high-contrast regions in individual scans, X-VORTEX integrates information across time to preserve coherent localization.

Qualitative performance is especially relevant in operational settings, where the most challenging cases arise not during early, high-contrast wake stages, but during later phases when vortices are partially dissipated yet still potentially hazardous. The consistency of X-VORTEX in these borderline conditions aligns with its design goal of learning representations robust to wake decay and sensor sparsity. Additional qualitative examples across diverse wake conditions are provided in Appendix~\ref{sec:visualizations}.

\section{Conclusion}
\label{sec:conclusion}

We introduced \textbf{X-VORTEX}, a self-supervised spatio-temporal contrastive learning framework for wake vortex analysis from LiDAR point cloud sequences. By leveraging \emph{Augmentation Overlap Theory} and treating temporal evolution as a natural augmentation, X-VORTEX learns transferable sequence-level representations from large-scale unlabeled data. Experiments on nearly one million scans demonstrate that X-VORTEX substantially reduces labeling requirements for vortex center localization and enables accurate short-horizon trajectory forecasting, consistently outperforming heuristic, image-based, and fully supervised baselines. Ablation studies further confirm the importance of temporal subsampling, spatial masking, sequence-level normalization, and explicit temporal modeling in achieving these gains.

Future work will explore extending X-VORTEX to longer-horizon forecasting with uncertainty estimation, integrating continuous-time temporal models, and incorporating meteorological context such as wind and turbulence profiles. Beyond wake vortices, we believe the proposed spatio-temporal self-supervised framework can be applied to other atmospheric flow phenomena captured by remote sensing, opening new opportunities for data-driven airspace safety and efficiency.

\bibliography{example_paper}

@techreport{NTSB_AA587,
  title       = {In-Flight Separation of Vertical Stabilizer, American Airlines Flight 587, Airbus Industrie A300-605R, N14053, Belle Harbor, New York, November 12, 2001},
  author      = {{National Transportation Safety Board}},
  year        = {2004},
  number      = {NTSB/AAR-04/04},
  institution = {National Transportation Safety Board},
  address     = {Washington, D.C.}
}

@techreport{BFU_Challenger,
  title       = {Interim Report: Serious Incident, Bombardier Challenger 604 vs Airbus A380-800, Arabian Sea},
  author      = {{Bundesstelle f{\"u}r Flugunfalluntersuchung (BFU)}},
  year        = {2017},
  month       = {May},
  number      = {BFU 17-0024-2X},
  institution = {German Federal Bureau of Aircraft Accident Investigation},
  address     = {Braunschweig, Germany}
}

@article{hallock2018review,
  title={A review of recent wake vortex research for increasing airport capacity},
  author={Hallock, James N and Holz{\"a}pfel, Frank},
  journal={Progress in Aerospace Sciences},
  volume={98},
  pages={27--36},
  year={2018},
  publisher={Elsevier}
}

@article{holzapfel2021mitigating,
  title={Mitigating wake turbulence risk during final approach via plate lines},
  author={Holz{\"a}pfel, Frank and Stephan, Anton and Rotshteyn, Grigory and K{\"o}rner, Stephan and Wildmann, Norman and Oswald, Lothar and Gerz, Thomas and Borek, G{\"u}nther and Floh, Alexander and Kern, Christian and others},
  journal={AIAA Journal},
  volume={59},
  number={11},
  pages={4626--4641},
  year={2021},
  publisher={American Institute of Aeronautics and Astronautics}
}

@inproceedings{qu2025explainable,
  title={Explainable LiDAR 3D Point Cloud Segmentation and Clustering for Detecting Airplane-Generated Wind Turbulence},
  author={Qu, Zhan and Yuan, Shuzhou and F{\"a}rber, Michael and Brennfleck, Marius and Wartha, Niklas and Stephan, Anton},
  booktitle={Proceedings of the 31st ACM SIGKDD Conference on Knowledge Discovery and Data Mining V. 1},
  pages={2504--2513},
  year={2025}
}

@misc{eurocontrol2018european,
  title = "{European Aviation in 2040. Challenges of Growth}",
  author={EUROCONTROL},
  year={2018},
  url={https://www.eurocontrol.int/sites/default/files/content/documents/official-documents/reports/challenges-of-growth-2018.pdf}
}

@article{frech2002short,
  title={Short-term prediction of the horizontal wind vector within a wake vortex warning system},
  author={Frech, Michael and Holz{\"a}pfel, Frank and Gerz, Thomas and Konopka, Jens},
  journal={Meteorological Applications},
  volume={9},
  number={1},
  pages={9--20},
  year={2002},
  publisher={Cambridge University Press}
}

@article{pan2022identification,
  title={Identification of aircraft wake vortex based on VGGNet},
  author={Pan, Weijun and Leng, Yuanfei and Yin, Haoran and Zhang, Xiaolei and others},
  journal={Wireless Communications and Mobile Computing},
  volume={2022},
  year={2022},
  publisher={Hindawi}
}

@article{robins2001algorithm,
  title={Algorithm for prediction of trailing vortex evolution},
  author={Robins, Robert E and Delisi, Donald P and Greene, George C},
  journal={Journal of aircraft},
  volume={38},
  number={5},
  pages={911--917},
  year={2001}
}

@article{korner2019assessment,
  title={Assessment of the wake-vortex proximity to landing aircraft exploiting field measurements},
  author={K{\"o}rner, Stephan and Holz{\"a}pfel, Frank},
  journal={Journal of Aircraft},
  volume={56},
  number={3},
  pages={1250--1258},
  year={2019},
  publisher={American Institute of Aeronautics and Astronautics}
}

@article{wartha2022characterizing,
  title={Characterizing aircraft wake vortex position and strength using LiDAR measurements processed with artificial neural networks},
  author={Wartha, Niklas and Stephan, Anton and Holz{\"a}pfel, Frank and Rotshteyn, Grigory},
  journal={Optics Express},
  volume={30},
  number={8},
  pages={13197--13225},
  year={2022},
  publisher={Optica Publishing Group}
}

@inproceedings{stephan2023artificial,
  title={Artificial Neural Networks for Individual Tracking and Characterization of Wake Vortices in LIDAR Measurements},
  author={Stephan, Anton and Rotshteyn, Grigory and Wartha, Niklas and Holz{\"a}pfel, Frank N and Petross, Nicolas and Stietz, Lars},
  booktitle={AIAA AVIATION 2023 Forum},
  pages={3682},
  year={2023}
}

@article{smalikho2015method,
  title={Method of radial velocities for the estimation of aircraft wake vortex parameters from data measured by coherent Doppler lidar},
  author={Smalikho, Igor N and Banakh, VA and Holz{\"a}pfel, Frank and Rahm, Stephan},
  journal={Optics Express},
  volume={23},
  number={19},
  pages={A1194--A1207},
  year={2015},
  publisher={Optical Society of America}
}

@inproceedings{wangchaos,
  title={Chaos is a Ladder: A New Theoretical Understanding of Contrastive Learning via Augmentation Overlap},
  author={Wang, Yifei and Zhang, Qi and Wang, Yisen and Yang, Jiansheng and Lin, Zhouchen},
  booktitle={International Conference on Learning Representations}, 
  year={2022}
}

@article{zhang2025augmentation,
  title={An Augmentation Overlap Theory of Contrastive Learning},
  author={Zhang, Qi and Wang, Yifei and Wang, Yisen},
  journal={Journal of Machine Learning Research},
  volume={26},
  number={228},
  pages={1--42},
  year={2025}
}

@article{jianbing2017survey,
  title={A survey of the scattering characteristics and detection of aircraft wake vortices},
  author={Li, Jianbing and Gao, Hang and Wang, Tao and Wang, Xuesong},
  journal={Journal of Radars},
  volume={6},
  number={6},
  pages={660--672},
  year={2017},
  publisher={Journal of Radars}
}

@book{KUNDU2002629,
title = "Fluid Mechanics",
publisher = "Academic Press",
edition = "2nd",
pages = "629 - 660",
year = "2002",
isbn = "978-0-12-178251-1",
author = "Pijush K. Kundu and Ira M. Cohen"
}

@article{stephan2019effects,
  title={Effects of detailed aircraft geometry on wake vortex dynamics during landing},
  author={Stephan, Anton and Rohlmann, David and Holz{\"a}pfel, Frank and Rudnik, Ralf},
  journal={Journal of Aircraft},
  volume={56},
  number={3},
  pages={974--989},
  year={2019},
  publisher={American Institute of Aeronautics and Astronautics}
}

@article{harvey1971flowfield,
  title={Flowfield produced by trailing vortices in the vicinity of the ground},
  author={Harvey, JK and Perry, Fi J},
  journal={AIAA journal},
  volume={9},
  number={8},
  pages={1659--1660},
  year={1971}
}

@article{spalart2001modeling,
  title={Modeling the interaction of a vortex pair with the ground},
  author={Spalart, PR and Strelets, M Kh and Travin, AK and Shur, ML},
  journal={Fluid Dynamics},
  volume={36},
  number={6},
  pages={899--908},
  year={2001},
  publisher={Springer}
}

@article{holzapfel2007aircraft,
  title={Aircraft wake-vortex evolution in ground proximity: analysis and parameterization},
  author={Holz{\"a}pfel, Frank and Steen, Meiko},
  journal={American Institute of Aeronautics and Astronautics journal},
  volume={45},
  number={1},
  pages={218--227},
  year={2007}
}

@article{zheng1996study,
  title={Study of aircraft wake vortex behavior near the ground},
  author={Zheng, ZC and Ash, Robert L},
  journal={AIAA journal},
  volume={34},
  number={3},
  pages={580--589},
  year={1996},
}

@article{pan2020identification,
  title={Identification of aircraft wake vortex based on SVM},
  author={Pan, Weijun and Wu, Zhengyuan and Zhang, Xiaolei},
  journal={Mathematical Problems in Engineering},
  volume={2020},
  pages={1--8},
  year={2020},
  publisher={Hindawi Limited}
}

@inproceedings{weijun2021aircraft,
  title={Aircraft wake detection method based on SVM and morphology},
  author={Pan, Weijun and Leng, Yuanfei and Han, Shuai},
  booktitle={2021 IEEE 3rd International Conference on Civil Aviation Safety and Information Technology (ICCASIT)},
  pages={356--360},
  year={2021},
  organization={IEEE}
}

@inproceedings{abras2021application,
  title={Application of machine learning to automate flow-physics identification in computed solutions: Hover rotor wake vortex identification and breakdown analysis},
  author={Abras, Jennifer and Hariharan, Nathan S},
  booktitle={AIAA Scitech 2021 Forum},
  pages={0474},
  year={2021}
}

@article{ai2021deep,
  title={A deep learning framework based on multisensor fusion information to identify the airplane wake vortex},
  author={Ai, Yi and Wang, Yuanji and Pan, Weijun and Wu, Dingjie},
  journal={Journal of Sensors},
  volume={2021},
  pages={1--10},
  year={2021},
  publisher={Hindawi Limited}
}

@article{chu2024assessment,
  title={Assessment of approach separation with probabilistic aircraft wake vortex recognition via deep learning},
  author={Chu, Nana and Ng, Kam KH and Liu, Ye and Hon, Kai Kwong and Chan, Pak Wai and Li, Jianbing and Zhang, Xiaoge},
  journal={Transportation Research Part E: Logistics and Transportation Review},
  volume={181},
  pages={103387},
  year={2024},
  publisher={Elsevier}
}

@article{pan2022recognition,
  title={Recognition of aircraft wake vortex based on random forest},
  author={Pan, Weijun and Yin, Haoran and Leng, Yuanfei and Zhang, Xiaolei},
  journal={IEEE Access},
  volume={10},
  pages={8916--8923},
  year={2022},
  publisher={IEEE}
}

@phdthesis{stietz2022artificial,
  title={Artificial Neural Networks for Individual Tracking and Characterization of Wake Vortices in LiDAR Measurements},
  author={Stietz, Lars Olaf},
  year={2022},
  school={Universit{\"a}t Hamburg}
}

@article{DGCNN_WSLSBS19,
  author       = {Yue Wang and
                  Yongbin Sun and
                  Ziwei Liu and
                  Sanjay E. Sarma and
                  Michael M. Bronstein and
                  Justin M. Solomon},
  title        = {Dynamic Graph {CNN} for Learning on Point Clouds},
  journal      = {{ACM} Trans. Graph.},
  volume       = {38},
  number       = {5},
  pages        = {146:1--146:12},
  year         = {2019}
}

@inproceedings{qi2017pointnet,
  title={Pointnet: Deep learning on point sets for 3d classification and segmentation},
  author={Qi, Charles R and Su, Hao and Mo, Kaichun and Guibas, Leonidas J},
  booktitle={Proceedings of the IEEE conference on computer vision and pattern recognition},
  pages={652--660},
  year={2017}
}

@article{qi2017pointnet++,
  title={Pointnet++: Deep hierarchical feature learning on point sets in a metric space},
  author={Qi, Charles Ruizhongtai and Yi, Li and Su, Hao and Guibas, Leonidas J},
  journal={Advances in neural information processing systems},
  volume={30},
  year={2017}
}

@article{Survey3D_GWHLLB21,
  author       = {Yulan Guo and
                  Hanyun Wang and
                  Qingyong Hu and
                  Hao Liu and
                  Li Liu and
                  Mohammed Bennamoun},
  title        = {Deep Learning for 3D Point Clouds: {A} Survey},
  journal      = {{IEEE} Trans. Pattern Anal. Mach. Intell.},
  volume       = {43},
  number       = {12},
  pages        = {4338--4364},
  year         = {2021}
}

@article{weijun2019deep,
  title={Deep learning for aircraft wake vortex identification},
  author={Pan, Weijun and Duan, Yingjie and Zhang, Qiang and Tang, Jiahao and Zhou, Jun},
  booktitle={IOP Conference Series: Materials Science and Engineering},
  journal = {IOP Conference Series: Materials Science and Engineering},
  publisher = {IOP Publishing},
  volume={685},
  number={1},
  pages={012015},
  year={2019},
  organization={IOP Publishing}
}

@inproceedings{choy20194d,
  title={4d spatio-temporal convnets: Minkowski convolutional neural networks},
  author={Choy, Christopher and Gwak, JunYoung and Savarese, Silvio},
  booktitle={Proceedings of the IEEE/CVF conference on computer vision and pattern recognition},
  pages={3075--3084},
  year={2019}
}

@inproceedings{aygun20214d,
  title={4d panoptic lidar segmentation},
  author={Aygun, Mehmet and Osep, Aljosa and Weber, Mark and Maximov, Maxim and Stachniss, Cyrill and Behley, Jens and Leal-Taix{\'e}, Laura},
  booktitle={Proceedings of the IEEE/CVF Conference on Computer Vision and Pattern Recognition},
  pages={5527--5537},
  year={2021}
}

@inproceedings{liu2019meteornet,
  title={Meteornet: Deep learning on dynamic 3d point cloud sequences},
  author={Liu, Xingyu and Yan, Mengyuan and Bohg, Jeannette},
  booktitle={Proceedings of the IEEE/CVF International Conference on Computer Vision},
  pages={9246--9255},
  year={2019}
}

@article{fan2022pstnet,
  title={Pstnet: Point spatio-temporal convolution on point cloud sequences},
  author={Fan, Hehe and Yu, Xin and Ding, Yuhang and Yang, Yi and Kankanhalli, Mohan},
  journal={arXiv preprint arXiv:2205.13713},
  year={2022}
}

@article{fan2019pointrnn,
  title={PointRNN: Point recurrent neural network for moving point cloud processing},
  author={Fan, Hehe and Yang, Yi},
  journal={arXiv preprint arXiv:1910.08287},
  year={2019}
}

@inproceedings{fan2021point,
  title={Point 4d transformer networks for spatio-temporal modeling in point cloud videos},
  author={Fan, Hehe and Yang, Yi and Kankanhalli, Mohan},
  booktitle={Proceedings of the IEEE/CVF conference on computer vision and pattern recognition},
  pages={14204--14213},
  year={2021}
}

@inproceedings{chen2020simple,
  title={A simple framework for contrastive learning of visual representations},
  author={Chen, Ting and Kornblith, Simon and Norouzi, Mohammad and Hinton, Geoffrey},
  booktitle={International conference on machine learning},
  pages={1597--1607},
  year={2020},
  organization={PmLR}
}

@inproceedings{he2020momentum,
  title={Momentum contrast for unsupervised visual representation learning},
  author={He, Kaiming and Fan, Haoqi and Wu, Yuxin and Xie, Saining and Girshick, Ross},
  booktitle={Proceedings of the IEEE/CVF conference on computer vision and pattern recognition},
  pages={9729--9738},
  year={2020}
}

@inproceedings{xie2020pointcontrast,
  title={Pointcontrast: Unsupervised pre-training for 3d point cloud understanding},
  author={Xie, Saining and Gu, Jiatao and Guo, Demi and Qi, Charles R and Guibas, Leonidas and Litany, Or},
  booktitle={European conference on computer vision},
  pages={574--591},
  year={2020},
  organization={Springer}
}

@inproceedings{zhang2021self,
  title={Self-supervised pretraining of 3d features on any point-cloud},
  author={Zhang, Zaiwei and Girdhar, Rohit and Joulin, Armand and Misra, Ishan},
  booktitle={Proceedings of the IEEE/CVF international conference on computer vision},
  pages={10252--10263},
  year={2021}
}

@inproceedings{wang2020understanding,
  title={Understanding contrastive representation learning through alignment and uniformity on the hypersphere},
  author={Wang, Tongzhou and Isola, Phillip},
  booktitle={International conference on machine learning},
  pages={9929--9939},
  year={2020},
  organization={PMLR}
}

@inproceedings{wartha2023investigating,
  title={Investigating Errors of Wake Vortex Retrievals Using High Fidelity Lidar Simulations},
  author={Wartha, Niklas and Stephan, Anton and Holz{\"a}pfel, Frank N and Rotshteyn, Grigory},
  booktitle={AIAA AVIATION 2023 Forum},
  pages={3679},
  year={2023},
}

@article{felton2025characterizing,
  title={Characterizing wake vortex pairs using nonsupervised machine learning},
  author={Felton, Melvin A and Kumar, Prabhat and Catuche, Jana and James, Deryck and Ligon, David A},
  journal={Optics Express},
  volume={33},
  number={18},
  pages={37795--37813},
  year={2025},
  publisher={Optica Publishing Group}
}

@article{zhang2024locating,
  title={Locating and Grading of Lidar-Observed Aircraft Wake Vortex Based on Convolutional Neural Networks},
  author={Zhang, Xinyu and Zhang, Hongwei and Wang, Qichao and Liu, Xiaoying and Liu, Shouxin and Zhang, Rongchuan and Li, Rongzhong and Wu, Songhua},
  journal={Remote Sensing},
  volume={16},
  number={8},
  pages={1463},
  year={2024},
  publisher={MDPI}
}

@article{baranov2021wake,
  title={Wake vortex detection by convolutional neural networks},
  author={Baranov, Nikolay and Resnick, Boris},
  journal={European Journal of Electrical Engineering and Computer Science (EEACS)},
  volume={3},
  pages={92--97},
  year={2021}
}
\bibliographystyle{icml2026}

%%%%%%%%%%%%%%%%%%%%%%%%%%%%%%%%%%%%%%%%%%%%%%%%%%%%%%%%%%%%%%%%%%%%%%%%%%%%%%%
%%%%%%%%%%%%%%%%%%%%%%%%%%%%%%%%%%%%%%%%%%%%%%%%%%%%%%%%%%%%%%%%%%%%%%%%%%%%%%%
% APPENDIX
%%%%%%%%%%%%%%%%%%%%%%%%%%%%%%%%%%%%%%%%%%%%%%%%%%%%%%%%%%%%%%%%%%%%%%%%%%%%%%%
%%%%%%%%%%%%%%%%%%%%%%%%%%%%%%%%%%%%%%%%%%%%%%%%%%%%%%%%%%%%%%%%%%%%%%%%%%%%%%%
\newpage
\appendix
\onecolumn
\section{Aircraft Classes and Linear Probing Setup}
\label{app:probing}

To assess representation quality, we perform linear probing using aircraft type classification as a diagnostic task. Although aircraft type prediction is not a downstream objective of interest, it serves as a proxy for wake structure variability induced by differences in aircraft geometry and mass. The probing dataset contains 16 aircraft classes. Table~\ref{tab:aircraft_counts} summarizes the class distribution.

\begin{table}[h]
\centering
\caption{Aircraft classes and sample counts used in the linear probing evaluation.}
\label{tab:aircraft_counts}
\resizebox{0.29\linewidth}{!}{%
\begin{tabular}{l r l r}
\toprule
Aircraft & Count & Aircraft & Count \\
\midrule
\texttt{A20N} & 714     & \texttt{B744} & 262 \\
\texttt{A310} & 69      & \texttt{B748} & 176 \\
\texttt{A320} & 20{,}337 & \texttt{B752} & 287 \\
\texttt{A332} & 198     & \texttt{B763} & 4{,}728 \\
\texttt{A333} & 374     & \texttt{B772} & 3{,}546 \\
\texttt{A359} & 183     & \texttt{B77L} & 393 \\
\texttt{A388} & 353     & \texttt{B77W} & 484 \\
              &         & \texttt{B788} & 462 \\
              &         & \texttt{B789} & 502 \\
\bottomrule
\end{tabular}}
\end{table}

The class distribution is highly imbalanced, with \texttt{A320}, \texttt{B763}, and \texttt{B772} dominating the dataset. Consequently, linear probing accuracy should be interpreted primarily as a relative indicator of representation quality rather than absolute classification performance. All probing experiments follow the labeled split described in Section~\ref{sec:experiments_data}, with the backbone frozen and a single linear classifier trained on top of the learned sequence-level embeddings.

\section{Alignment and Uniformity}
\label{sec:align_uniform}

To better study the behavior of X-VORTEX during self-supervised pre-training, we analyze the learned embedding geometry using the \textbf{alignment} and \textbf{uniformity} metrics introduced by Wang and Isola~\cite{wang2020understanding}. Alignment measures the expected distance between positive pairs, reflecting how tightly augmented views of the same sequence are pulled together. Uniformity measures how evenly embeddings are distributed over the unit hypersphere, penalizing representation collapse and promoting separation between different samples.

\begin{figure}[!h]
    \centering
    \includegraphics[width=0.49\linewidth]{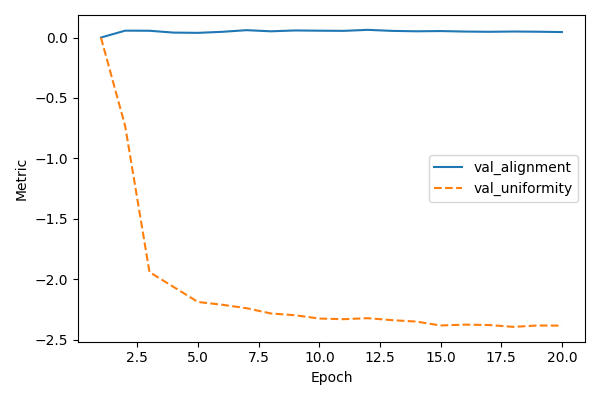}
    \caption{\textbf{Alignment and uniformity during self-supervised pre-training (PointNet backbone).}
    Validation alignment and uniformity over the first 20 epochs are shown for clarity. Alignment improves rapidly as positive pairs are pulled together, while uniformity decreases steadily, indicating increasing dispersion of embeddings across the hypersphere. Both metrics stabilize early, suggesting stable contrastive training without representation collapse.}
    \label{fig:align_uniform}
\end{figure}

Figure~\ref{fig:align_uniform} reports validation alignment and uniformity over the first 20 pre-training epochs for the PointNet backbone (shown for clarity). Alignment rapidly improves within the first few epochs and then stabilizes, indicating that temporally and spatially augmented views of the same wake sequence are consistently mapped to nearby embeddings. At the same time, uniformity decreases steadily (becoming more negative), reflecting improved dispersion of representations across the embedding space.

Concretely, alignment increases from near zero at initialization to stable values around $0.05$, while uniformity drops from approximately $0$ to about $-2.38$, suggesting that X-VORTEX simultaneously achieves strong intra-sequence invariance and maintains inter-sequence diversity. Importantly, both metrics stabilize early in training, indicating well-behaved contrastive optimization without embedding collapse.

This balance between alignment and uniformity helps explain the strong downstream performance observed with limited labeled data: tight alignment encourages robustness to temporal subsampling and spatial sparsity, while good uniformity preserves discriminability across different wake events. Together, these properties support effective transfer of the learned representations to localization and forecasting tasks.

\section{Additional Qualitative Visualizations}
\label{sec:visualizations}

This appendix provides additional qualitative examples for both wake vortex center localization and trajectory forecasting, complementing the representative cases shown in Sec.~\ref{sec:qualitative_results}. Each visualization displays Doppler LiDAR radial velocity fields, where warm colors indicate motion toward the sensor and cool colors indicate motion away. Green and orange circles denote ground-truth centers of the port and starboard vortices, respectively, while crosses indicate model predictions.

\begin{figure*}[!hb]
    \centering
    \includegraphics[width=\textwidth]{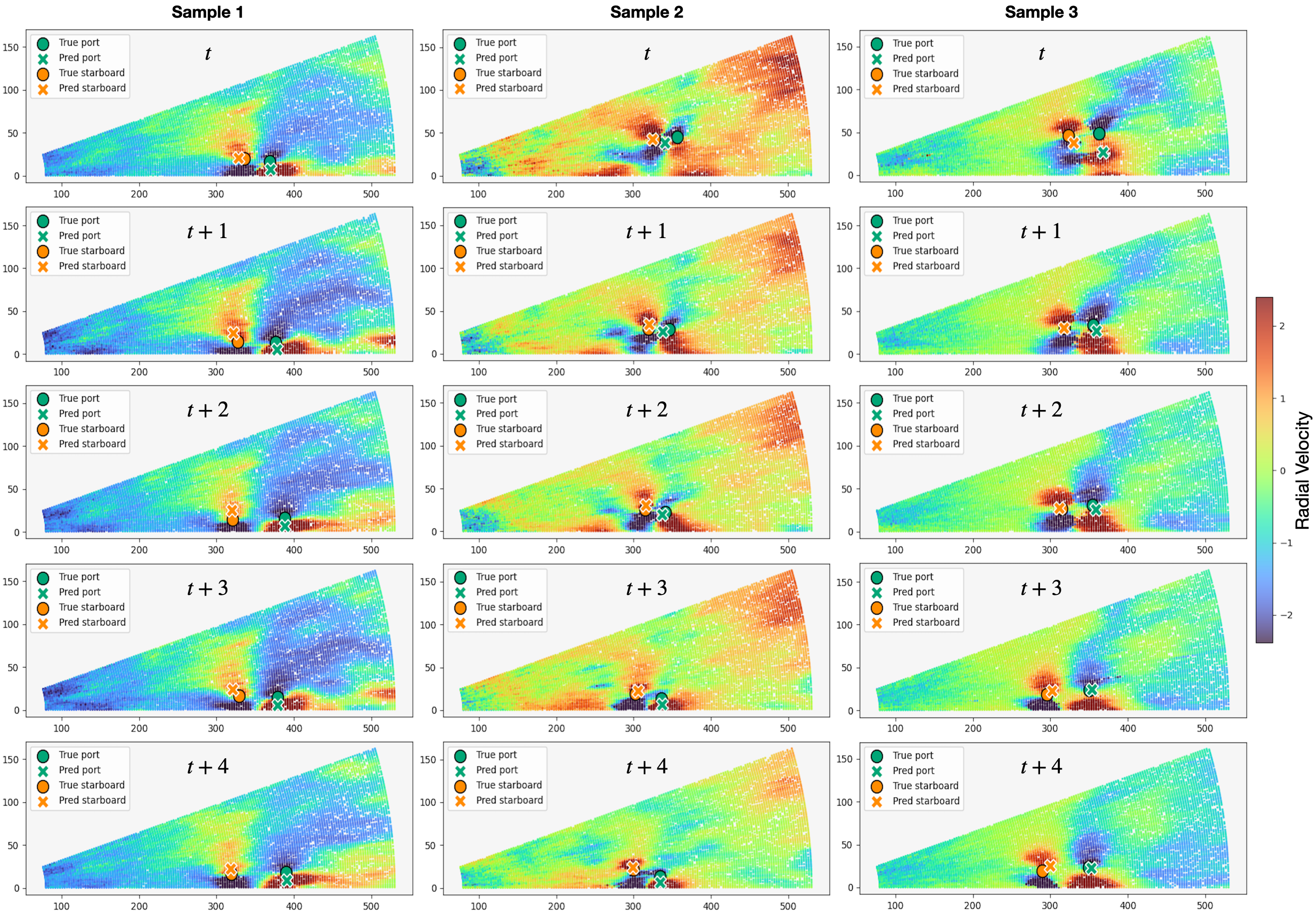}
    \caption{\textbf{Additional center localization examples (Samples 1--3).}
    Each column shows a wake sequence over five consecutive scans ($t$ to $t{+}4$). Circles indicate ground-truth vortex centers and crosses denote X-VORTEX predictions.}
    \label{fig:appendix_localization_1}
\end{figure*}

\begin{figure*}[!ht]
    \centering
    \includegraphics[width=\textwidth]{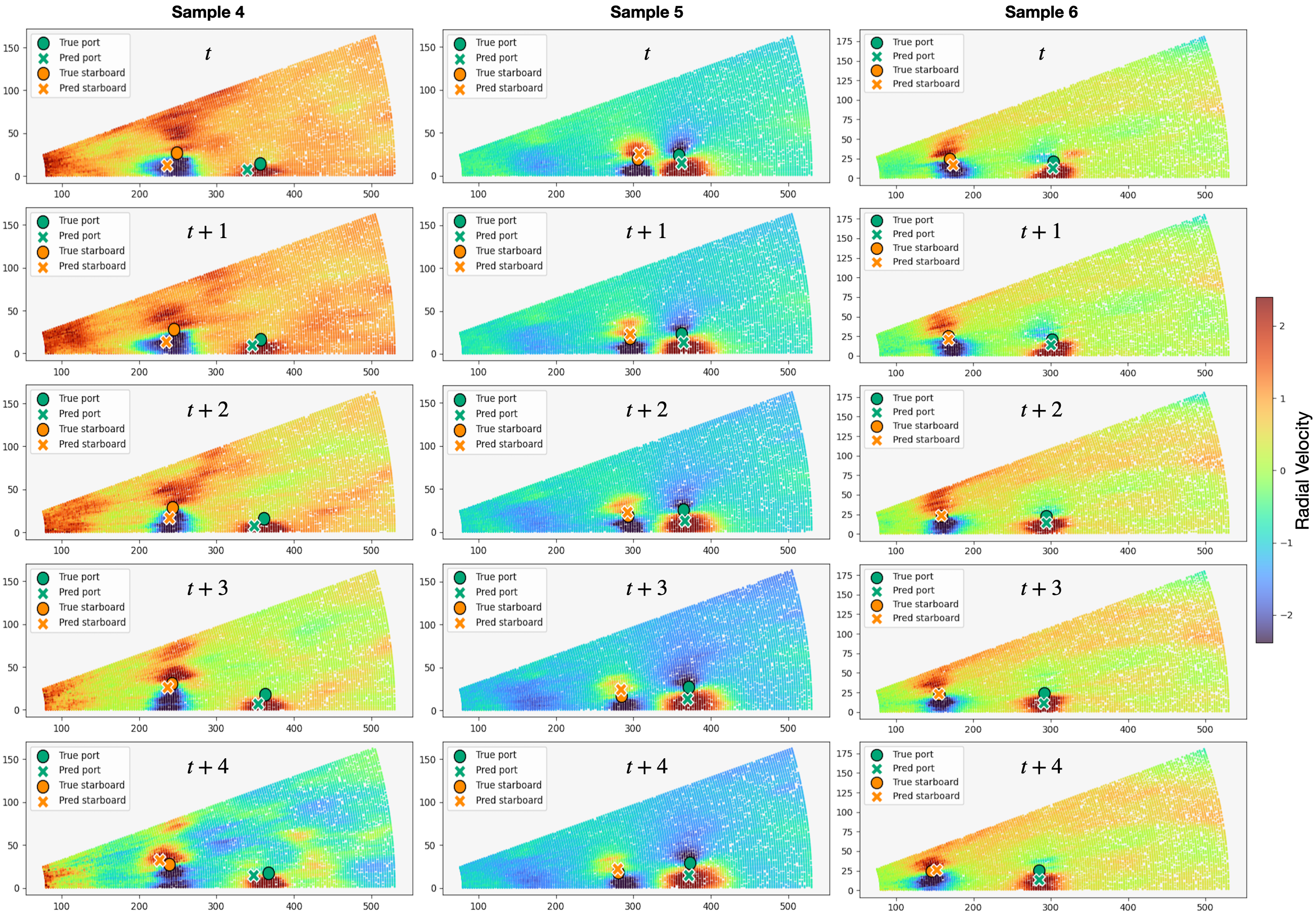}
    \caption{\textbf{Additional center localization examples (Samples 4--6).}
    X-VORTEX maintains consistent localization across diverse wake conditions and progressive decay.}
    \label{fig:appendix_localization_2}
\end{figure*}

\paragraph{Center Localization.}
Figures~\ref{fig:appendix_localization_1} and~\ref{fig:appendix_localization_2} show two sets of three test sequences, each consisting of five consecutive scans ($t$ to $t{+}4$). These examples cover diverse wake strengths, decay patterns, and background conditions. Despite progressive weakening of velocity contrast and increasing sparsity in later frames, X-VORTEX consistently tracks both vortex cores and maintains stable predictions across time.

\begin{figure*}[!ht]
    \centering
    \includegraphics[width=\textwidth]{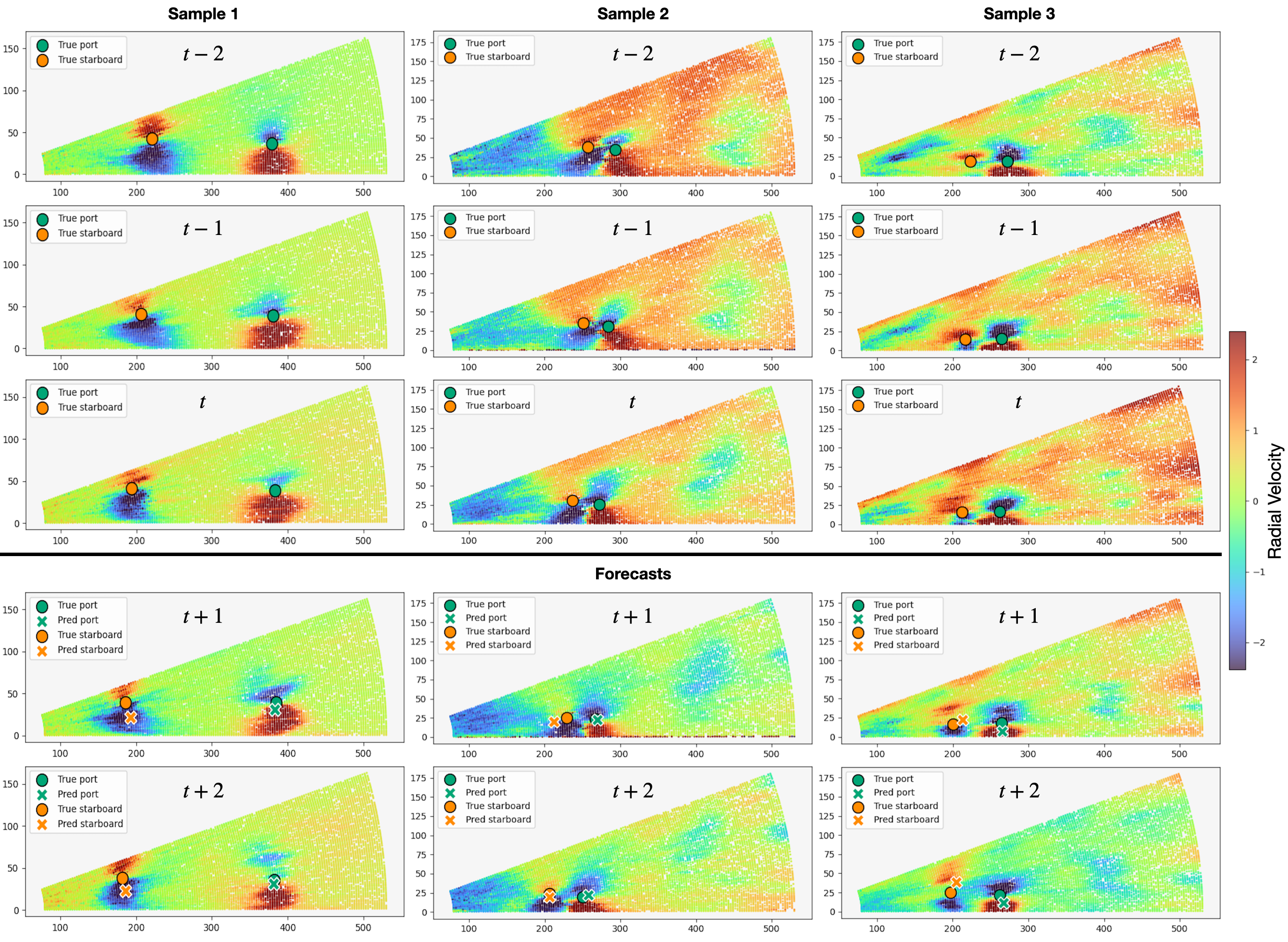}
    \caption{\textbf{Additional trajectory forecasting examples.}
    Ground-truth centers are shown up to $t$, followed by X-VORTEX predictions at $t{+}1$ and $t{+}2$, illustrating short-horizon extrapolation under non-linear wake evolution.}
    \label{fig:appendix_forecasting}
\end{figure*}

\paragraph{Trajectory Forecasting.}
Figure~\ref{fig:appendix_forecasting} presents three representative forecasting examples. For each sequence, ground-truth centers are shown up to $t$, followed by model predictions at $t{+}1$ and $t{+}2$. These cases illustrate X-VORTEX’s ability to extrapolate wake motion beyond the observed history, even under non-linear descent and asymmetric decay between the port and starboard vortices.

Across all samples, X-VORTEX demonstrates robustness to wake dissipation, point sparsity, and background clutter, supporting the quantitative improvements reported in Secs.~\ref{sec:label_efficiency} and~\ref{sec:forecasting}.

%%%%%%%%%%%%%%%%%%%%%%%%%%%%%%%%%%%%%%%%%%%%%%%%%%%%%%%%%%%%%%%%%%%%%%%%%%%%%%%
%%%%%%%%%%%%%%%%%%%%%%%%%%%%%%%%%%%%%%%%%%%%%%%%%%%%%%%%%%%%%%%%%%%%%%%%%%%%%%%

\end{document}